%% file: preprint.tex
\documentclass{article}

\usepackage{fullpage}
\usepackage{amsmath}
\usepackage{amsfonts}
\usepackage{tikz}
\tikzset{>=latex}
\usepackage{xcolor}

\title{On Training Recurrent Networks with Truncated Backpropagation Through Time
    in Speech Recognition}

\author{Hao Tang, James Glass}
\date{Computer Science and Artificial Intelligence Laboratory \\
  Massachusetts Institute of Technology \\
  Cambridge, MA 02139, USA \\
  {\small\texttt{\{haotang,glass\}@mit.edu}}}

\newcommand{\textfw}[1]{{\small\texttt{#1}}}

\begin{document}

\maketitle

\input{abstract.tex}

\input{intro.tex}

\input{prob.tex}

\input{bptt.tex}

\input{exp.tex}

\input{markov.tex}

\input{conclusion.tex}

\bibliographystyle{plain}
\bibliography{tbptt}

\end{document}

%% file: abstract.tex
\begin{abstract}
Recurrent neural networks have been the dominant models for many speech
and language processing tasks. However, we understand
little about the behavior and the class of functions recurrent networks can
realize. Moreover, the heuristics used during training complicate the
analyses. In this paper, we study recurrent networks' ability to learn
long-term dependency in the context of speech recognition. We consider
two decoding approaches, online and batch decoding, and show the classes
of functions to which the decoding approaches correspond. We then draw
a connection between batch decoding and a popular training approach
for recurrent networks, truncated backpropagation through time. Changing
the decoding approach restricts the amount of past history recurrent
networks can use for prediction, allowing us to analyze their ability to
remember. Empirically, we utilize long-term dependency in subphonetic
states, phonemes, and words, and show how the design decisions, such
as the decoding approach, lookahead, context frames, and consecutive
prediction, characterize the behavior of recurrent networks. Finally, we
draw a connection between Markov processes and vanishing gradients. These
results have implications for studying the long-term dependency in speech
data and how these properties are learned by recurrent networks.
\end{abstract}

%% file: intro.tex
\section{Introduction}

Recurrent neural networks (RNNs) have been extensively used for speech
and language processing, and are particularly successful at
sequence prediction tasks,
such as language modeling \cite{MKS2018-1,MKS2018-2,KHQJ2018}
and automatic speech recognition \cite{SSEP2014,SSB2014,PWPK2018}.
Their success is often attributed to their ability to learn long-term dependencies
in the data or, more generally, their ability to remember.
However, exactly how far back recurrent networks can remember \cite{WLS2004, KHQJ2018},
whether there is a limit \cite{CSS2017}, and how the limit is characterized
by the network architecture~\cite{Z+2016} or by the optimization method ~\cite{PMB2013},
are still open questions.
In this paper, we examine how training and decoding approaches affect
recurrent networks' ability to learn long-term dependencies.
In particular, 
we study the behavior of recurrent networks in the context of speech recognition
when their ability to remember is constrained.

The ability to remember in recurrent networks
is affected by many factors, such as the amount of long-term dependency
in the data, as well as the training and decoding approaches.
We differentiate between two types of decoding, the online approach
and the batch approach.  In the online case,
there is a single chain of recurrent network that predicts
at all time steps, while in the batch case,
multiple chains of recurrent networks that start
and end at different time points are used.
In the online case, a recurrent network resembles a recursive function
where the prediction of the current time step depends on
the memory that encodes the entire history.  In the batch
case, a recurrent network resembles a fixed order Markov process,
because the prediction at the current time step strictly depends
on a fixed number of time steps in the past.
By changing the decoding approach, we restrict
the ability of recurrent networks to model certain classes of functions.

The speech signal is particularly rich with long-term dependencies.
Due to the causal nature of time, speech is assumed to be Markovian
(though with a potentially high order).
The order of the Markov property depends on the time scale
and the linguistic units, such as subphonetic states, phonemes, and words.
The Markov assumption has a strong influence in many design choices,
such as the model family or the training approaches.
Partly due to the Markov assumption and partly due to
computational efficiency \cite{KTLK2015,CYH2015}, recurrent networks
in speech recognition \cite{SSEP2014,SSB2014}
and in language modeling \cite{MKS2018-1, MKS2018-2, KHQJ2018} are commonly
trained with truncated backpropagation through time (BPTT) \cite{W1990, WP1990},
where a recurrent network is unfolded for a fixed number of time steps
for the gradients to propagate.
The hypothesis is that even with truncation,
recurrent networks are still able to learn recursive
functions that are richer than fixed order Markov processes.
In fact, recurrent networks at test time are typically applied to sequences much
longer than the ones they are trained on~\cite{SSB2014,PWPK2018}.
Under certain conditions, it has be shown that
truncation does not affect the performance of RNNs \cite{MH2018}.
By our definition, these recurrent networks are trained in batch mode,
and are used in an online fashion during testing.
We will examine how the number of BPTT steps affects
the training loss and how the decoding approaches
at test time affect the test loss.

Another factor that impacts the ability of a recurrent network 
to remember is optimization \cite{BSF1994, PMB2013}.
Vanilla recurrent networks are known to be difficult
to train \cite{HBFS2001,PMB2013}.
Previous work has attributed this difficulty to the
vanishing gradient problem \cite{H1998}.
Long short-term memory networks (LSTMs)
have been proposed to alleviate the vanishing gradient problem and are
assumed to be able to learn longer dependencies in the data \cite{HS1997}.
Whether these statements are true is still debatable, but we will draw a connection between the
Markov assumption, Lipschitz continuity, and vanishing
gradients in recurrent networks.  The vanishing gradient phenomenon
is in fact a necessary condition for Markov processes.

The contribution of the paper is a comprehensive set of
experiments comparing two decoding approaches, various numbers of BPTT steps during
training, and their respective results on three types of
target units, namely, subphonetic states, phonemes, and words.
The results have implications for studying long-term dependency
in speech and how well they are learned by recurrent networks.

%% file: prob.tex
\section{Problem Setting}
\label{sec:prob}

We first review the definition of recurrent networks
and discuss the types of functions they aim to model.

Let $X$ be the set of input elements, $L$ be the label set,
and $Y = \mathbb{R}^{|L|}$ for representing
(log-)probabilities or one-hot vectors of elements in $L$.
For example, in the case of speech recognition, the set $X = \mathbb{R}^{40}$
if we use 40-dimensional acoustic features, and $L$ can be the set of
subphonetic states, phonemes, or words.
Let $\mathcal{X}, \mathcal{Y}$ be the sets of sequences whose elements are in
$X$, $Y$, respectively.
We are interested in finding a function that maps a sequence in $\mathcal{X}$
to a sequence in $\mathcal{Y}$ of the same length.
In other words, the goal is to map $(x_1, \dots, x_T) \in \mathcal{X}$,
where each $x_i \in X$, to $(y_1, \dots, y_T) \in \mathcal{Y}$,
where each $y_i \in Y$, for $i = 1, \dots, T$.
We assume we have access to a loss function
$\ell: Y \times Y \to \mathbb{R}^+$.
For example, the loss function for classification at each time step $t$
can be the cross entropy $\ell(y_t, \hat{y}_t) = - y_t^\top \log \hat{y}_t$,
where $y_t$ is the one-hot vector
for the ground truth label at time $t$, and $\log \hat{y}_t$
is a vector of log probabilities produced by the model.

A general recurrent network is a function $s: H \times X \to H$
where $H$ is the space of hidden vectors.
For an input sequence $(x_1, \dots, x_T)$, we repeatedly apply
\begin{align}
h_t = s(h_{t-1}, x_t)
\end{align}
for $t = 1, \dots, T$
to obtain a sequence of hidden vectors $(h_1, \dots, h_T)$
where $h_0 = 0$.
The hidden vectors are then used for prediction.
Note that we are interested in the setting where the length of the
input sequence matches the length of the output sequence.
Specifically, we have a function $o: H \to Y$ and apply
\begin{align}
\hat{y}_t = o(h_t)
\end{align}
for $t = 1, \dots, T$ to obtain the label sequence $(\hat{y}_1, \dots, \hat{y}_T)$.
For our purposes, it suffices to use $o(h_t) = \text{softmax}(W h_t)$
for some trainable parameter matrix $W$.

A vanilla recurrent neural network implements the above with
\begin{align}
s(h_{t-1}, x_t) & = \sigma(U x_t + V h_{t-1}),
\end{align}
while a long short-term memory network (LSTM) \cite{HS1997} uses
\begin{align}
\begin{bmatrix}
g_t \\
i_t \\
f_t \\
o_t
\end{bmatrix}
& =
\begin{pmatrix}
\tanh \\
\sigma \\
\sigma \\
\sigma
\end{pmatrix}
(U x_t + V h_{t-1}) \\
c_t &= i_t \odot g_t + f_t \odot c_{t-1} \\
s(h_{t-1}, x_t) &= o_t \odot \tanh(c_t)
\end{align}
to implement the recurrent network, where
$\odot$ is the Hadamard product,
$\sigma$ is the logistic function, $\tanh$
is the hyperbolic tangent, and the matrices
$U$ and $V$ are trainable parameters.

Recurrent networks can benefit from stacking on top of each other~\cite{GMH2013}.
We abstract away stacking in the discussion, but will use
stacked recurrent networks in the experiments.

\subsection{Markov and recursive functions}
\label{sec:markov}

Recurrent networks can be seen as recursive functions
with a constant size memory.
To be precise, consider $y_t$ as a function of $x_1, \dots, x_t$ for any $t$.
We say that $y_t$ is a recursive function if it satisfies
$y_t = f(m_t)$ where $m_i = g(m_{i-1}, x_i)$
for $i = 1, \dots, t$ and any function $f$ and $g$.
The memory is of size constant if the size of $m_t$
is independent of $t$.
In contrast, we say that $y_t$ is a $\kappa$-th order Markov function if 
$y_t$ is a function of $x_{t-\kappa+1}, \dots, x_t$.
By this definition, the set of recursive functions
includes all Markov functions, and the inclusion is proper.
For example, the sum $y_t = \sum_{i=1}^t x_i$
is not Markov of a fixed order, but it can be realized with
a recursive function $y_t = m_t$ where $m_i = m_{i-1} + x_i$
for $i = 1, \dots, t$.
In the language of signal processing,
a Markov function resembles a finite-impulse response filter,
and a recursive function resembles an infinite-impulse response
filter.

\subsection{Online and batch decoding}

Based on the two function classes,
we define two decoding approaches, online and batch
decoding, respectively.  In online decoding, a sequence of
predictions is made one after another,
while in batch decoding, the predictions
at different time steps are made independently
from each other.

Using the notation in the previous section,
to predict $\hat{y}_t$ at time $t$, we define online decoding as
$\hat{y}_t = f(m_t)$ where $m_i = g(m_{i-1}, x_i)$ for $i = 1, \dots, t$
and any function $f$ and $g$.
During decoding, only the vector $m_i$ at time $i$
needs to be stored in memory, and no history is actually maintained;
hence the name online decoding.
To implement online decoding with recurrent networks,
we simply let $f = o$, $g = s$, and $m_i = h_i$ for all $i$.

We define batch decoding as $\hat{y}_t = f(x_{t-\kappa+1}, \dots, x_t)$
for some function $f$ and context size $\kappa$.
By this definition, batch decoding is not limited to recurrent
networks, and can be used with other neural networks.
To implement batch decoding with recurrent networks,
we let $f$ compute $\hat{y}_t = o(h_t)$, $h_t = z^t_t$, and
$z^t_i = s(z^t_{i-1}, x_i)$ where $z^t_{t - \kappa} = 0$,
for $i = t - \kappa + 1, \dots, t$.

In terms of computation graphs, the graph for online
decoding with recurrent networks is a single chain, while the graph
for batch decoding consists of multiple parallel chains.
Note that in batch decoding the hidden vector of each chain
starts from the zero vector. In other words, $h_i$ is not a function of $h_{i-1}$
for any $i$, so the hidden vectors cannot be reused when
predicting at different time points.
Though batch decoding requires more computation and space,
it can be parallelized.  Online decoding has to be computed
sequentially.

By the above definition, batch decoding with $\kappa$ context frames aims to
realize a $\kappa$-th order Markov function.
Online decoding, however, aims to realize a recursive function.
It has been shown that the number of unrolled steps in
vanilla recurrent networks controls the capacity of the model
in terms of the Vapnik-Chervonenkis dimension \cite{KS1996}.
It is unclear whether recurrent networks truly learn this class of functions.
However, by changing decoding approaches, we explicitly restrict
the class of functions that recurrent networks can realize.

Finally, one distinct property in speech recognition
that is absent in language modeling is the option to
look beyond the current time frame.  
The task of predicting the next word becomes meaningless
when the next word is observed.
However, it is useful to look a few frames
ahead to predict the word while the word is being spoken.
Formally, we say that a recurrent
network decodes with a lookahead $\ell$ if $\hat{y}_t = o(h_{t + \ell - 1})$.
Lookahead can be applied to both online and batch decoding,
and it falls back to regular decoding when the lookahead is one frame.

%% file: bptt.tex
\section{Backpropagation through time}

There are many approaches to training recurrent networks~\cite{A1987,P1987},
and the most successful one by far is backpropagation
through time \cite{W1990}.
Backpropagation through time (BPTT) is an approach to compute
the gradient of a recurrent network when consuming a sequence.
A computation graph is created based on the decoding approaches,
and the gradients are propagated back through the computation
vertices.  It is equivalent to unrolling the recurrent networks
for several time steps depending on the decoding approaches;
hence the name backpropagation through time.

There are many variants of BPTT with the most popular one
being truncated BPTT \cite{WP1990}.  In the original
definition \cite{WP1990}, instead of propagating
gradients all the way to the start of the unrolled chain,
the accumulation stops after a fixed number of time steps.
Training recurrent networks with truncated BPTT can be justified
if the truncated chains are enough to learn
the target recursive functions.
In the modern setting~\cite{SSEP2014},
truncated BPTT is treated as regular BPTT with batch decoding,
and the number of unrolled steps before truncation is
the number of context frames in batch decoding.
Henceforth, we will use the term BPTT with batch decoding
to avoid confusion.

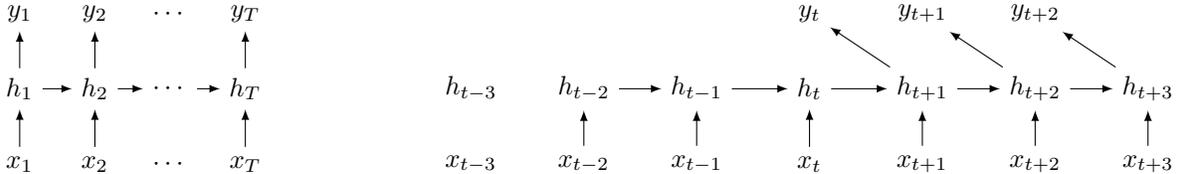
\begin{figure*}
\begin{center}
\begin{tikzpicture}
\node (x1) at (0, 0) {$x_1$} ;
\node (x2) at (1, 0) {$x_2$} ;
\node (xdots) at (2, 0) {$\cdots$} ;
\node (xT) at (3, 0) {$x_T$} ;

\node (h1) at (0, 1) {$h_1$} ;
\node (h2) at (1, 1) {$h_2$} ;
\node (hdots) at (2, 1) {$\cdots$} ;
\node (hT) at (3, 1) {$h_T$} ;

\node (y1) at (0, 2) {$y_1$} ;
\node (y2) at (1, 2) {$y_2$} ;
\node (ydots) at (2, 2) {$\cdots$} ;
\node (yT) at (3, 2) {$y_T$} ;

\draw[->] (x1) -- (h1);
\draw[->] (x2) -- (h2);
\draw[->] (xT) -- (hT);

\draw[->] (h1) -- (h2);
\draw[->] (h2) -- (hdots);
\draw[->] (hdots) -- (hT);

\draw[->] (h1) -- (y1);
\draw[->] (h2) -- (y2);
\draw[->] (hT) -- (yT);

\node (x2T3) at (6, 0) {$x_{t-3}$} ;
\node (x2T2) at (7.5, 0) {$x_{t-2}$} ;
\node (x2T1) at (9, 0) {$x_{t-1}$} ;
\node (x2T) at (10.5, 0) {$x_t$} ;
\node (x2Tp1) at (12, 0) {$x_{t+1}$} ;
\node (x2Tp2) at (13.5, 0) {$x_{t+2}$} ;
\node (x2Tp3) at (15, 0) {$x_{t+3}$} ;

\node (h2T3) at (6, 1) {$h_{t-3}$} ;
\node (h2T2) at (7.5, 1) {$h_{t-2}$} ;
\node (h2T1) at (9, 1) {$h_{t-1}$} ;
\node (h2T) at (10.5, 1) {$h_t$} ;
\node (h2Tp1) at (12, 1) {$h_{t+1}$} ;
\node (h2Tp2) at (13.5, 1) {$h_{t+2}$} ;
\node (h2Tp3) at (15, 1) {$h_{t+3}$} ;

\node (y2T) at (10.5, 2) {$y_t$} ;
\node (y2Tp1) at (12, 2) {$y_{t+1}$} ;
\node (y2Tp2) at (13.5, 2) {$y_{t+2}$} ;

\draw[->] (h2T2) -- (h2T1);
\draw[->] (h2T1) -- (h2T);
\draw[->] (h2T) -- (h2Tp1);
\draw[->] (h2Tp1) -- (h2Tp2);
\draw[->] (h2Tp2) -- (h2Tp3);

\draw[->] (x2T2) -- (h2T2);
\draw[->] (x2T1) -- (h2T1);
\draw[->] (x2T) -- (h2T);
\draw[->] (x2Tp1) -- (h2Tp1);
\draw[->] (x2Tp2) -- (h2Tp2);
\draw[->] (x2Tp3) -- (h2Tp3);

\draw[->] (h2Tp1) -- (y2T);
\draw[->] (h2Tp2) -- (y2Tp1);
\draw[->] (h2Tp3) -- (y2Tp2);

\end{tikzpicture}
\caption{\emph{Left}: Online decoding for $(x_1, \dots, x_T)$.
    \emph{Right}: One chain of batch decoding
    at time $t$ with 6 context frames, a lookahead of 2 frames, and consecutive prediction
    of 3 frames.  The chains are repeated for $t = 1, \dots, T$.}
\label{fig:online-batch}
\end{center}
\vspace{-0.5cm}
\end{figure*}

BPTT with batch decoding is seldom used in practice
due to the high computation cost,
with the only exception being~\cite{SSEP2014}
where the context is set to six frames.
A more common approach used in conjunction with BPTT and batch decoding
is to predict a batch of frames rather than a single one \cite{SSB2014, CYH2015}.
Formally, for a single chain of recurrent network to
predict $p$ consecutive frames at time $t$, we have
$\hat{y}_{t+i-1} = o(h_{t+\ell+i-2})$ for $i = 1, \dots, p$
where $\ell$ is the number of lookahead.
Note that this decoding approach is only used for training
and is never actually used at test time.
Figure~\ref{fig:online-batch} summarizes
the decoding approaches and the hyperparameters, including
lookahead, context frames, and consecutive prediction.

In practice, recurrent networks are trained with a combination
of BPTT, batch decoding, lookahead, and consecutive prediction \cite{SSB2014, CYH2015}.
To speed up training, no frames in an utterance are predicted more than once,
and many sequences are processed in batches to better
utilize parallel computation on GPUs.
In addition, the hidden vectors are sometimes cached \cite{SSEP2014}.
Applying these heuristics, however, creates a mismatch
between training and testing.
Previous work has not addressed this issue,
and the distinction between online and batch decoding
under BPTT has only been lightly explored in~\cite{SSEP2014, KTLK2015}.
We will examine these in detail in our experiments.

%% file: exp.tex
\section{Experiments}

To study how well recurrent networks model Markov
and recursive functions, we conduct experiments on
frame classification tasks with three different
linguistic units, i.e., subphonetic states, phonemes,
and words.  We expect that subphonetic states and phonemes are
relatively local in time, while words require
longer contexts to predict.  By varying the target units, we control
the amount of long-term dependency in the data.

Experiments are conducted on the Wall Street Journal data set
(\textfw{WSJ0} and \textfw{WSJ1}), consisting of about 80 hours of
read speech and about 20,000 unique words,
suitable for studying rich long-term dependency.
We use 90\% of \textfw{si284}
for training, 10\% of \textfw{si284} for development,
and evaluate the models on \textfw{dev93}.
The set \textfw{dev93} is chosen because
it is the only set where frame error rates are
reported for deep networks \cite{GJM2013}.
The time-aligned frame labels, including
subphonetic states, phonemes, and words,
are obtained from speaker-adaptive hidden Markov models
following the Kaldi recipe~\cite{P+2011}.

In the following experiments, we use, as input to the frame classifiers,
80-dimensional log Mel features without appending i-vectors.
For recurrent networks, we use 3-layer LSTMs with 512 hidden units
in each layer.  In addition, for the baseline we use
a 7-layer time-delay neural network (TDNN)
with 512 hidden units in each layer and an architecture
described in~\cite{THGG2018,PWPK2018}.
The network parameters are initialized based on~\cite{HZRS2015}.
The networks are trained with vanilla stochastic gradient descent
with step size 0.05 for 20 epochs.
The batch size is one utterance and the gradients are clipped
to norm 5.  The best performing model is chosen based on
the frame error rates (FER) on the development set.
Utterances are padded with zeros when lookahead or
context frames outside the boundaries is queried.

\subsection{Long-term dependency in subphonetic states}

\begin{figure}[t]
\begin{center}
\includegraphics[width=5cm]{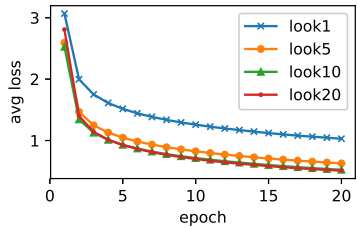}
\caption{Running average of training losses
    for comparing different amount of lookahead for LSTMs
    with online decoding trained on subphonetic states.}
\label{fig:train}
\end{center}
\vspace{-0.5cm}
\end{figure}

We first experiment with recurrent networks on subphonetic states.
We expect LSTMs to perform best when trained and
tested with online decoding, so we first explore
the effect of lookahead.
Results are shown in Table~\ref{tbl:state-lookahead}.
Lookahead has a significant effect on the frame error rates.
The error rate improves as we increase the amount of lookahead
and plateaus after 10 frames.
The improvement is due to better training error (not regularization or other factors),
as shown in Figure~\ref{fig:train}.
In addition, the fact that increasing the amount of lookahead does
not hurt performance suggests that LSTMs are able
to retain information for at least 20 frames.
Compared to prior work, our 3-layer LSTM is unidirectional
and uses fewer layers than the ones used in~\cite{GJM2013},
but the results are on par with theirs.
The best LSTM achieves 11.7\% word error rate on \textfw{dev93},
similar to the results in~\cite{GJM2013}.\footnote{
We are aware of the state of the art on this data set \cite{CL2015}.
As the results are in the ballpark,
we do not optimize them further.}
The TDNN serves as an instance of a Markov function.
We present its result, but more investigations are needed
to conclude anything from it.

\begin{table}[t]
\caption{FERs (\%) comparing different LSTM lookaheads
    trained with online decoding, on subphonetic states.
    The FER of a 7-layer TDNN is provided as a reference.}
\label{tbl:state-lookahead}
\begin{center}
\begin{tabular}{lll}
lookahead & dev       & \textfw{dev93}   \\
\hline 
1         & 45.46     & 43.45   \\
5         & 33.64     & 32.43   \\
10        & 30.43     & 29.83   \\
15        & 29.89     & 28.69   \\
20        & 29.36     & 28.59   \\
\hline
\hline
TDNN 512x7
          & 33.56     & 34.19
\end{tabular}
\end{center}
\end{table}

To see how decoding approaches affect performance,
we examine the error rates with batch decoding 
on the best LSTM in Table~\ref{tbl:state-lookahead},
i.e., the one trained with online decoding
and a lookahead of 20 frames.  Results are shown in
Table~\ref{tbl:state-batch}.  The performance
deteriorates as we reduce the number of context frames.
This suggests that LSTMs do utilize information
40 frames away.  On the other hand, the degradation
is not severe, suggesting that much of
the long-term dependency is Markov.

\begin{table}[t]
\caption{FERs (\%) comparing different context frames
    under batch decoding
    with the best LSTM trained with online decoding,
    and a lookahead of 20 frames on subphonetic states.}
\label{tbl:state-batch}
\begin{center}
\begin{tabular}{lll}
context & dev       & \textfw{dev93}   \\
\hline
40      & 37.65     & 36.45   \\
35      & 41.73     & 40.12   \\
30      & 48.79     & 46.70   \\
\hline
\hline
online  & 29.36     & 28.59
\end{tabular}
\end{center}
\vspace{-0.5cm}
\end{table}

We then examine whether LSTMs with
batch decoding can learn recursive functions.
The same LSTMs are trained with batch decoding
of different context frames.
Consecutive prediction is not used in this setting.
In other words, for an utterance of $T$ frames,
we create $T$ chains of LSTMs,
each of which predicts the label of a frame.
Results are shown in Table~\ref{tbl:state-tbptt}.
With a context of 40 frames, the LSTM trained
with batch decoding is only slightly behind the LSTM trained
with online decoding.  This again suggests
that the class of Markov functions can perform reasonably well,
and much of the long-term dependency is likely Markov.
However, the error rate degrades significantly
when we switch from batch decoding to online decoding.
This strongly suggests that these LSTMs do not
behave like recursive functions.

\begin{table}[t]
\caption{FERs (\%) on \textfw{dev93} comparing online and batch decoding
    for LSTMs trained with batch decoding,
    and a lookahead of 20 frames on subphonetic states.}
\label{tbl:state-tbptt}
\begin{center}
\begin{tabular}{lll}
context & batch   & online \\
\hline
40      & 31.30   & 80.29  \\
35      & 30.99   & 84.78  \\
30      & 32.80   & 85.74       
\end{tabular}
\end{center}
\vspace{-0.5cm}
\end{table}

To understand what makes LSTMs
behave like recursive functions,
we train LSTMs with batch decoding
and increase the amount of consecutive
prediction.  To simplify the setting,
we allow each frame to be predicted multiple
times.  In fact, if a network predicts
consecutively for $p$ frames,
then each frame gets predicted $p$ times.
Results are shown in Table~\ref{tbl:state-pred}.
As we increase the number of consecutively predicted frames,
the error rate for batch decoding
stays about the same, while the one for online
decoding improves.  This suggests that
it is the amount of consecutive prediction that gears
the behavior of LSTMs towards recursive functions.
In addition, as seen in Table~\ref{tbl:state-batch},
LSTMs can achieve reasonable performance
with both online and batch decoding.
Perhaps the data is a complex mix of long-term
dependency, or perhaps the LSTM learns to
forget the history.  More analyses are needed
to tease the factors apart.

\begin{table}[t]
\caption{FERs (\%) on \textfw{dev93} comparing different
    numbers of consecutive prediction
    for LSTMs trained with batch decoding,
    and a lookahead of 20 frames on subphonetic states.}
\label{tbl:state-pred}
\begin{center}
\begin{tabular}{llll}
\# of prediction & batch   & online \\
\hline 
1                & 31.30   & 80.29  \\
5                & 31.29   & 80.21  \\
10               & 31.65   & 46.82  \\
15               & 32.27   & 33.03  
\end{tabular}
\end{center}
\vspace{-0.5cm}
\end{table}

\subsection{Long-term dependency in phonemes and words}

We repeat the same experiments with phonemes and words.
Note that the number of classes in the case of words
is the number of unique words in the training set.
We do not handle out-of-vocabulary words (OOV).

For the lookahead experiments in Table~\ref{tbl:phone-word-lookahead},
the general trend is similar to that in Table~\ref{tbl:state-lookahead},
except that the frame error rate plateaus at around 15 frames
for words, longer than the 10 frames in subphonetic states
and phonemes.

The fact that the TDNN achieves
a low frame error rate for phonemes in Table~\ref{tbl:phone-word-lookahead}
suggests that much phonetic information
is concentrated within a 30-frame window.
However, for words we expect a larger context
window to predict words, so we evaluate a 10-layer
TDNN with an effective context window of 48 frames.
The 10-layer TDNN performs better than the 7-layer
one, but is still behind LSTMs.

The conclusion in Table~\ref{tbl:phone-word-batch}
is the same as in Table~\ref{tbl:state-batch},
i.e., for LSTMs trained with online decoding,
reducing the amount of context during batch decoding
hurts the accuracy.  However, the conclusion
in Table~\ref{tbl:phone-word-tbptt} is different
from that in Table~\ref{tbl:state-tbptt}.
The LSTMs with batch decoding are
able to perform well with online decoding
on phonemes but not on words.
We suspect that phonemes are more local than subphonetic states,
and in general the results are affected by the choice of
linguistic units.

In Table~\ref{tbl:phone-word-pred}, we observe a similar trend to
that in Table~\ref{tbl:state-pred}, adding consecutive prediction
improves the LSTMs' performance with online decoding.
In fact, the performance with online decoding,
compared to that in Table~\ref{tbl:phone-word-lookahead},
is fully recovered when using 15 frames of consecutive prediction.


\begin{table}[t]
\caption{FERs (\%) comparing different lookaheads
    for LSTMs trained with online decoding, on phonemes and words.
    The FERs of a 7-layer TDNN and a 10-layer TDNN are provided as reference.}
\label{tbl:phone-word-lookahead}
\begin{center}
\begin{tabular}{l|ll|ll}
          & \multicolumn{2}{c}{phonemes} & \multicolumn{2}{|c}{words} \\
lookahead & dev    & \textfw{dev93}  & dev    & \textfw{dev93}   \\
\hline                                                          
1         & 18.31  & 16.34           & 32.78  & 45.60            \\
5         & 13.57  & 12.01           & 28.39  & 40.36            \\
10        & 12.33  & 11.01           & 24.68  & 36.62            \\
15        & 11.90  & 10.67           & 21.57  & 32.82            \\
20        & 11.86  & 10.76           & 20.85  & 31.45            \\
25        &        &                 & 21.02  & 30.43            \\
\hline
\hline
TDNN 512x7
          & 14.22  & 12.46           & 39.27  & 38.93            \\
TDNN 512x10
          &        &                 & 28.97  & 32.24            
\end{tabular}
\end{center}
\vspace{-0.5cm}
\end{table}

\begin{table}[t]
\caption{FERs (\%) for batch decoding with different context frames
    with the best LSTM trained with online decoding,
    and a lookahead of 20 frames on phonemes and words.}
\label{tbl:phone-word-batch}
\begin{center}
\begin{tabular}{l|ll|ll}
        & \multicolumn{2}{c}{phonemes} & \multicolumn{2}{|c}{words} \\
context & dev   & \textfw{dev93} & dev   & \textfw{dev93} \\
\hline                                                   
40      & 16.74 & 14.99          & 48.39 & 44.84          \\
35      & 19.28 & 17.10          & 53.33 & 48.55          \\
30      & 23.26 & 20.59          & 58.78 & 53.00          \\
\hline                                                    
\hline                                                    
online  & 11.86 & 10.76          & 20.85 & 31.45          
\end{tabular}
\end{center}
\vspace{-0.5cm}
\end{table}

\begin{table}[t]
\caption{FERs (\%) on \textfw{dev93} comparing online and batch decoding
    for LSTMs trained with batch decoding,
    and a lookahead of 20 frames on phonemes and words.}
\label{tbl:phone-word-tbptt}
\begin{center}
\begin{tabular}{l|ll|ll}
         & \multicolumn{2}{c}{phonemes} & \multicolumn{2}{|c}{words} \\
context  & batch & online & batch & online \\
\hline                   
40       & 11.50 & 11.24  & 31.65 & 56.96  \\
35       & 12.12 & 15.11  & 34.21 & 58.26  \\
30       & 12.89 & 21.97  & 37.83 & 64.74  
\end{tabular}
\end{center}
\vspace{-0.5cm}
\end{table}

\begin{table}[t]
\caption{FERs (\%) on \textfw{dev93} comparing different
    numbers of consecutive prediction
    for LSTMs trained with batch decoding,
    and a lookahead of 20 frames on phonemes and words.}
\label{tbl:phone-word-pred}
\begin{center}
\begin{tabular}{l|ll|ll}
                 & \multicolumn{2}{c}{phonemes} & \multicolumn{2}{|c}{words} \\
\# of prediction & batch & online & batch & online \\
\hline                            
1                & 11.50 & 11.24  & 31.65 & 56.96  \\
5                & 11.58 & 11.68  & 31.79 & 45.88  \\
10               & 11.49 & 10.89  & 32.31 & 36.40  \\
15               & 11.48 & 10.72  & 32.34 & 30.93  
\end{tabular}
\end{center}
\vspace{-0.5cm}
\end{table}

%% file: markov.tex
\section{Markov Assumption, Lipschitz Continuity, and Vanishing Gradients}

From the experiments, it is difficult to conclude
whether the LSTMs really learn Markov or recursive functions.
In this section, we derive a necessary condition
that we can check empirically for Markov functions.
A function $f$ is $G$-Lipschitz with respect to $i$-th coordinate and norm $\|\cdot\|$ if
\begin{align}
& \bigg|f(a_1, \dots, a_{i-1}, a, a_{i+1}, \dots, a_t) \notag \\
& \quad - f(a_1, \dots, a_{i-1}, b, a_{i+1}, \dots, a_t)\bigg| \leq G\|a - b\|
\end{align}
for any $a_1, \dots, a_t$ and any $a$, $b$.
In words, the function does not change too much when we perturb
the $i$-th coordinate.
By definition, a Markov function does not change at all
when we perturb the coordinate beyond the necessary history,
or more formally, a $\kappa$-th order Markov function is 0-Lipschitz with
respect to $x_1, \dots, x_{t-\kappa}$.
The Lipschitz property relates to the gradient through
the perturbation interpretation.
A gradient of a function can be regarded as having
an infinitesimal perturbation of the coordinates,
so the $i$-th coordinate of a gradient has a value at most $G$
for a $G$-Lipschitz function with respect to the $i$-th coordinate.
A $\kappa$-th order Markov function should have zero gradient
with respect to $x_1, \dots, x_{t-\kappa}$,
because it is 0-Lipschitz with respect to those coordinates.
In other words, the gradients to the input, and thus to the parameters,
must vanish after $\kappa$ steps for a $\kappa$-th order Markov function.

\begin{figure}
\begin{center}
\includegraphics[width=5cm]{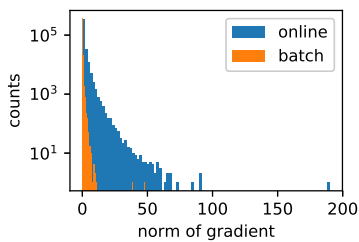}
\caption{The norms of gradients to the input $x_{t-20}$ when predicting $y_t$
    on \textfw{dev93}
    comparing LSTMs trained with online and batch decoding with 40 context frames.
    Both LSTMs are trained on subphonetic states with a lookahead of 20 frames.
    Note that the counts are in log scale.}
\label{fig:grad-hist}
\end{center}
\vspace{-0.5cm}
\end{figure}

In practice, partly because of noise and partly
because we do not know if the data
is really Markov, it is better not to enforce vanishing
gradients in the model.
Regardless, this gives a necessary condition we can check empirically.
We take the best online and batch LSTMs trained on subphonetic states
from Tables~\ref{tbl:state-lookahead} and \ref{tbl:state-tbptt}.
The norms of gradients to the input at time $t-20$
when predicting at time $t$
is shown in Figure~\ref{fig:grad-hist}.
The norms of the LSTM trained with batch decoding concentrate near
zero compared to the ones with online decoding.
This confirms that LSTMs trained with batch decoding behave
like Markov functions.

%% file: conclusion.tex
\section{Conclusion}

We study unidirectional recurrent networks, LSTMs in particular,
trained with two decoding approaches, online and batch
decoding, and explore various hyperparameters,
such as lookahead, context frames, and consecutive prediction.
Online decoding can be as broad as recursive functions,
while batch decoding can only include Markov functions.
Training LSTMs with the matching decoding approaches
performs best.  LSTMs trained with online decoding
can still have decent performance with batch decoding,
but LSTMs trained with batch decoding tend
not to perform well with online decoding.
The amount of lookahead is also critical
for LSTMs with online decoding to get better
training errors.  The number of context frames
strictly limits the history that can be used
for prediction, while increasing the amount of consecutive
prediction gears LSTMs closer to recursive functions.
The results depend on the long-term dependency in the data,
and we confirm this by exploring subphonetic states, phonemes, and words.
Finally, we show that the vanishing gradient phenomenon
is a necessary condition for Markov functions
with respect to variables beyond the necessary history.
It is important to note that the same LSTM architecture
and the same number of model parameters can
have drastically different results and behaviors.
We hope the results improve the understanding
of recurrent networks, what they learn from
the data, and ultimately the understanding
of speech data itself.